\pdfoutput=1

\documentclass[11pt]{article}

\usepackage[]{acl}

\usepackage{times}
\usepackage{latexsym}

\usepackage[T1]{fontenc}
\newcommand{\tablestyle}[2]{\setlength{\tabcolsep}{#1}\renewcommand{\arraystretch}{#2}\centering\footnotesize}
\newlength\savewidth\newcommand\shline{\noalign{\global\savewidth\arrayrulewidth
		\global\arrayrulewidth .8pt}\hline\noalign{\global\arrayrulewidth\savewidth}}		
\newcommand\scline[1]{\noalign{\global\savewidth\arrayrulewidth
		\global\arrayrulewidth .8pt}\cline{#1}\noalign{\global\arrayrulewidth\savewidth}}

\usepackage[T1]{fontenc}

\usepackage[utf8]{inputenc}

\usepackage{microtype}

\usepackage{graphicx}
\usepackage{multirow}
\usepackage[most]{tcolorbox}
\newcommand{\napi}{4 }
\newcommand{\nhf}{6 }
\usepackage[capitalize,noabbrev]{cleveref}

\title{Ada-LEval: Evaluating long-context LLMs \\ with length-adaptable benchmarks}

\author{Chonghua Wang$^{2}$\thanks{\> The work was done during an internship at Shanghai AI Laboratory; $^{\dagger}$ Project Lead; $^{\ddagger}$ Corresponding Author. }, Haodong Duan$^{1 \dagger}$, Songyang Zhang$^{1}$, Dahua Lin$^{1}$, Kai Chen$^{1 \ddagger}$ \\
  $^{1}$Shanghai AI Laboratory \\
  $^{2}$Shanghai Jiao Tong University \\
  \texttt{philipwang@sjtu.edu.cn} \\
  \texttt{duanhaodong@pjlab.org.cn}
  }

\begin{document}
\maketitle

\begin{abstract}

Recently, the large language model (LLM) community has shown increasing interest in enhancing LLMs' capability to handle extremely long documents. 
As various long-text techniques and model architectures emerge, 
the precise and detailed evaluation of models' long-text capabilities has become increasingly important. 
Existing long-text evaluation benchmarks, such as L-Eval and LongBench, 
construct long-text test sets based on open-source datasets, 
focusing mainly on QA and summarization tasks. 
These datasets include test samples of varying lengths (from 2k to 32k+) entangled together, 
making it challenging to assess model capabilities across different length ranges. 
Moreover, they do not cover the ultralong settings (100k+ tokens) that the latest LLMs claim to achieve. 
In this paper, we introduce Ada-LEval, a length-adaptable benchmark for evaluating the long-context understanding of LLMs. 
Ada-LEval includes two challenging subsets, TSort and BestAnswer, 
which enable a more reliable evaluation of LLMs' long context capabilities. 
These benchmarks support intricate manipulation of the length of test cases, 
and can easily produce text samples up to 128k tokens. 
We evaluate \napi state-of-the-art closed-source API models and \nhf open-source models with Ada-LEval. 
The evaluation results demonstrate the limitations of current LLMs, especially in ultra-long-context settings.
Our code is available at \url{https://github.com/open-compass/Ada-LEval}. 

\end{abstract}
\section{Introduction}

\begin{figure}[ht]
    \centering
    \includegraphics[width=\linewidth]{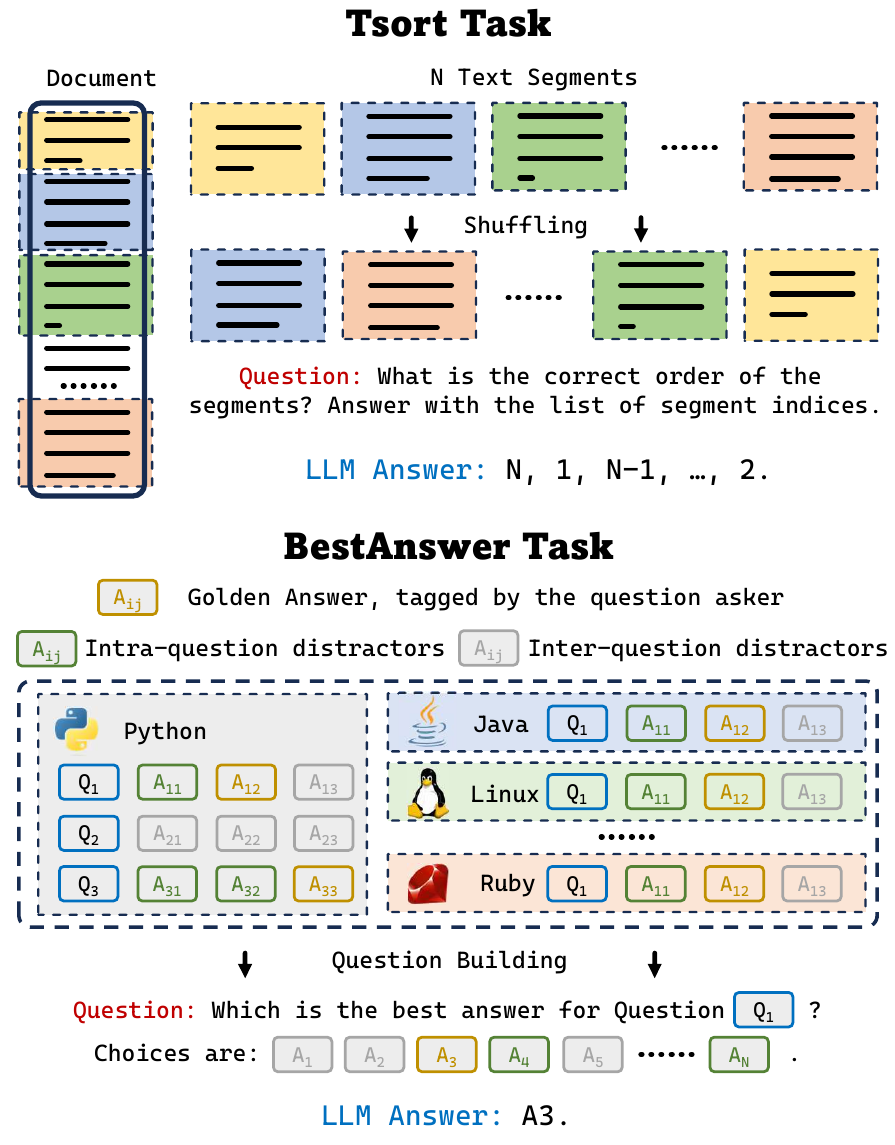}
    \caption{The demonstration of two tasks: \textbf{TSort} and \textbf{BestAnswer} introduced in Ada-LEval. 
    Understanding and reasoning over the full text are required to solve these two tasks.}
    \label{fig:adal-demo}
\end{figure}

Large Language Models (LLMs), typically based on large transformers trained on vast corpus, 
have shown exceptional abilities in memorization, comprehension, and reasoning~\citep{openai2023gpt4,touvron2023llama,zheng2023judging}. 
A critical factor that affects LLM performance is the `context window' - the number of tokens an LLM can process simultaneously. 
This window's size is pivotal in handling lengthy texts.
Since the debut of ChatGPT with a 2,000-token window in November 2022, 
significant efforts have been made in this domain, 
including more efficient attention mechanisms \citep{dao2022flashattention,zaheer2020big,ding2023longnet}, 
scalable position embeddings~\citep{su2021roformer,sun2022length}, 
and quantization techniques~\citep{frantar2022gptq,dettmers2022llm}.
As of December 2023, several LLMs claim to achieve context windows up to hundreds of thousands of tokens.
This includes both proprietary models like GPT-4 Turbo (128,000 tokens), Claude-2.1 (200,000 tokens), and Moonshot AI (200,000 Chinese characters), 
and open-source models such as ChatGLM-32k~\citep{zeng2022glm} and LongChat-32k~\citep{longchat2023}.
This expansion significantly enhances the potential for processing extensive documents. 
Nevertheless, the effectiveness of these long-context LLMs in managing long texts remains an area ripe for exploration and assessment.

Alongside the evolution of LLMs,
a wide range of benchmarks have emerged for capability assessment
\citep{hendrycks2020measuring,suzgun2022challenging,cobbe2021training,huang2023c}.
Most of those benchmarks utilize short questions or instructions, 
making them unsuitable for evaluating LLMs' long-context capabilities.
While a few benchmarks do focus on assessing specific long-context abilities 
like summarization, question-answering (QA), and continue writing
\citep{huang2021efficient,liu2023repobench,dasigi2021dataset},
comprehensive long-document evaluations have been limited. 
Recent benchmarks such as SCROLLS~\citep{shaham2022scrolls}, 
L-Eval~\citep{an2023eval} and LongBench~\citep{bai2023longbench} have started to address this gap by including a suite of long-document tasks, 
aiming for a more holistic assessment of LLMs' long-context understanding.

Despite these advancements, three significant limitations persist in existing benchmarks: 
Firstly, the ultra-long setting (32,000 tokens or longer) is scarcely represented, 
limiting insights into LLM performance in extreme context lengths. 
Secondly, the integration of test samples of varying lengths within these benchmarks complicates the evaluation of LLMs across different length ranges. 
Lastly, the focus on traditional tasks such as question-answering and summarization often does not necessitate comprehensive content understanding by the LLMs, 
as many questions in these tasks do not require full-text comprehension. 
This highlights the need for more targeted benchmarks that can rigorously evaluate the deep and complete understanding of long-form content by LLMs.

To this end, we introduce \textbf{Ada-LEval}, 
a pioneering benchmark to assess the long-context capabilities with length-adaptable questions.
Ada-LEval comprises two challenging tasks: \textbf{TSort}, which involves arranging text segments in the correct order, and \textbf{BestAnswer}, which requires choosing the best answer of a question among multiple candidates.  
Both tasks feature the following advantages:
1. \textbf{Controllable Test Cases}: The length of each test case can be finely tuned - 
by adjusting the number and length of text segments in TSort and 
altering the number of distractor options in BestAnswer. 
2. \textbf{Necessity for Full-Text Comprehension}: 
Successful completion of both tasks mandates complete reading and understanding of the provided text.
3. \textbf{Precise Accuracy Measurement}: The design of these tasks allows for unambiguous accuracy calculation. 
TSort has a definitive `correct' order, whereas in BestAnswer, the annotated responses by the questioner serve as definitive answers.

Our experiments on these tasks reveal critical insights. 
We observe a noteworthy decline in the performance of existing LLMs as text length increases, 
particularly in ultra-long scenarios. 
Furthermore, our ablation study uncovers several shortcomings in current LLMs, 
including limited instruction following over extended texts and
pronounced input order bias. 
Additionally, we explore various scalable position embedding techniques aimed at enlarging the context window of LLMs. 
Our findings indicate that models equipped with those techniques show improved performance over the standard models, 
and the performance is comparable to their counterparts trained on longer contexts.

\section{Related Work}
\subsection{Long-Context Techniques}

To address the complexities introduced by the increased text length in language models, researchers have developed a range of innovative techniques. These methodologies primarily focus on the following key areas:
more efficient attention mechanisms, divide-and-conquer paradigms, and scalable position embedding techniques. 

\noindent
\textbf{Efficient Attention Mechanisms. }
Notable advancements in attention mechanisms within Transformers have been achieved by several studies \citep{zaheer2020big, guo2021longt5, dao2022hungry, ding2023longnet}. 
A key development in this area is Flash Attention~\citep{dao2022flashattention},
which streamlines the attention process by circumventing the need to read and write the attention matrix across different memory tiers. 
This approach results in faster processing and reduced memory usage compared to traditional attention methods. 
In LongNet, \citet{ding2023longnet} introduces Dilated Attention, which reduces the computation complexity of attention to nearly linear and scales to 1 billion tokens. 
However, \citet{liu2023lost} identified a limitation where these mechanisms tend to falter with the middle portions of long texts.

\noindent
\textbf{Divide-and-Conquer. }
In exploring alternatives to conventional long-text modeling,
several studies have adopted a segmented approach to manage extensive content.
WebGPT~\citep{nakano2021webgpt} addresses long-form QA by interacting with a text-based web-browsing environment. 
PEARL~\citep{sun2023pearl} introduces a framework that prompts LLMs to generate and execute plans for tackling complex long-text reasoning tasks. 
\citet{chen2023walking} constructs a memory tree with the summarization of document segments and navigates on the memory tree to answer the original question.

\noindent
\textbf{Scalable Position Embeddings. }
Scalable position embeddings have been instrumental in extending the context window of LLMs.
RoPE~\citep{su2021roformer} utilizes a rotation matrix to enhance positional information, integrating explicit relative position dependencies into the self-attention mechanism. 
ALiBi~\citep{press2021train} does not add position embeddings to word embeddings, instead applying a linearly decreasing penalty to attention scores based on key-query distances.
Position Interpolation~\citep{chen2023extending} adopts a different strategy by linearly scaling down input position indices to align with preset context window sizes, requiring few fine-tuning steps.
NTK-aware Scaled RoPE\footnote{\url{https://www.reddit.com/r/LocalLLaMA/comments/14lz7j5/ntkaware_scaled_rope_allows_llama_models_to_have/}} and ReRoPE~\citep{rerope2023} further combine the benefits of position interpolation and length extrapolation methods without any fine-tuning steps.

\subsection{Long-Context Language Models}

Building on advancements in long-context techniques, 
several long-context LLMs are developed and released. 
Llama 2~\citep{touvron2023llama} integrates RoPE to expand its context window to 4,000 tokens. 
Vicuna-v1.5~\citep{zheng2023judging} further extends this capability by fine-tuning Llama 2 on high-quality, extensive conversations, successfully increasing the context window to 16,000 tokens. 
Longchat~\citep{longchat2023} models condense RoPE to utilize model weights learned in the pretraining stage. 
ChatGLM2-32k~\citep{zeng2022glm} is trained on a 32,000-token context length using position interpolation, showcasing the scalability of this technique. 

The domain of proprietary language models has seen even more significant advancements in long-context modeling, stepped into the ultra-long context field. 
GPT-4-Turbo~\cite{openai2023gpt4} notably extends its context window to an impressive 128,000 tokens. 
In a similar vein, Claude-2 and Claude-2.1 have achieved context lengths of 100,000 and 200,000 tokens respectively.
This expansion allows them to process vast quantities of information, such as hundreds of pages of technical documentation or entire books. 
Kimi Chat, developed by Moonshot.ai, claims to handle up to 200,000 Chinese characters.
However, no existing dataset can evaluate the capability in tackling such long texts.

\subsection{Long-Context Benchmarks}

Efforts to evaluate the long-context capabilities of language models have been intensifying,
with a focus primarily on traditional question-answering (QA) and summarization tasks.
NarrativeQA~\citep{kovcisky2018narrativeqa} offers a question-answering dataset built on the entire books from Project Gutenberg and movie transcripts. GovReport~\citep{huang2021efficient} provides a dataset comprising national policy issues, each accompanied by an expert-written summary, 
thus testing models' ability to distill complex, lengthy documents into concise summaries. 
Based on existing long-context benchmarks, 
SCROLLS\citep{shaham2022scrolls} introduces a suite of datasets that requires models to process and reason over long contexts.

Concurrently, L-Eval~\citep{an2023eval} and LongBench~\citep{bai2023longbench} are designed for comprehensive evaluation of long-context capabilities of LLMs. 
L-Eval offers a collection of long documents across different domains and provides both close-ended and open-ended tasks. 
LongBench is a bilingual long context benchmark covering six task categories. 
Most tasks in these benchmarks are traditional QA and summarization with fixed document, 
questions and answers. 
They are inflexible on text length (up to $\sim$32,000 tokens), which fall short of adapting to ultra-long context evaluation. 
Additionally, LongBench uses mostly open-ended tasks with traditional F1 and ROUGE metric that may not align well with human judgments. 
In contrast, our benchmarks support length-adaptable evaluation, provide sufficient cases and evaluate models using accuracy metrics, avoiding inconsistencies with human evaluation. 

\section{Ada-LEval}
In this section, we outline the construction process of Ada-LEval, 
detailing both the collection methodology of our source data and the building procedure of our test cases. 
\Cref{tab:statistics} demonstrates the data statistics of Ada-LEval.

\begin{table}[h]
    \centering
    \resizebox{1.0\linewidth}{!}{
        \begin{tabular}{c|c|c|c}
        \shline
        \multicolumn{4}{c}{\textbf{TSort}} \\ \shline
        Setting & Total \#Cases Built & Max \#Tokens & Avg \#Tokens \\
        \shline
        2k & 5123 & 2000 & 1816 \\
        4k & 5451 & 4000 & 3724 \\
        8k & 5324 & 8000 & 7663 \\
        16k & 4957 & 16000 & 15662 \\
        32k & 2206 & 32000 & 31226 \\
        64k & 1658 & 64000 & 62407 \\
        128k & 782 & 127800 & 121488 \\
        \shline
        \multicolumn{4}{c}{\textbf{BestAnswer}} \\ \shline
        Setting & Total \#Cases Built & Max \#Tokens & Avg \#Tokens \\ \shline
        1k & 7526 & 1128 & 955 \\
        2k & 7526 & 2154 & 1983 \\
        4k & 7526 & 4215 & 3994 \\
        6k & 7526 & 6268 & 6012 \\
        8k & 7526 & 7790 & 7518 \\
        12k & 7526 & 12389 & 12091 \\
        16k & 7526 & 15964 & 15646 \\
        32k & 200 & 32974 & 32329 \\
        64k & 200 & 64216 & 63274 \\
        128k & 200 & 127059 & 126098 \\
        \shline
        \end{tabular}
    }
    \caption{The data statistics of TSort and BestAnswer. 
    We adopt the GPT-4 tokenizer CL100K to calculate token numbers. 
    We use a subset of all built cases for evaluation. }
    \label{tab:statistics}
\end{table}

\subsection{Task Definition}
\textbf{TSort. } 
TSort provides LLMs with $\mathbf{N}$ shuffled text segments, extracted from contiguous chapters of a long book.
The task for models is to sort these segments into their original sequence. 
A response is regarded accurate only if it precisely reinstates the segments' initial order.
To simplify the challenge and minimize possible confusion, we supply LLMs with adjacent paragraphs from before and after the specified chapters to serve as contextual hints.

\noindent
\textbf{BestAnswer. } 
Each test case in BestAnswer contains one question and a large amount of possible answers to this question. 
We consider the answer designated by the original inquirer as the most helpful answer,
while LLMs are required to identify this optimal answer among all possible candidates.  

\subsection{Source Data Collection}

\textbf{TSort}. 
For TSort, we sourced our initial data from Booksum~\citep{kryscinski2021booksum}, 
a text summarization dataset derived from the Project Gutenberg, 
a public book repository consisting of over 60,000 free eBooks spanning various literary genres including novels, plays, short stories, and more. 
Genres like epistolary literature and poetry are excluded in the construction of TSort benchmark due to their non-sequential nature. 
To prevent LLMs from exploiting superficial cues, 
we meticulously remove identifiers such as chapter numbers and annotations from the content.

\noindent
\textbf{BestAnswer. } 
The BestAnswer benchmark is constructed using threads from Stack Overflow, a platform renowned for its extensive range of programming-related questions and answers. 
Stack Overflow questions are categorized by multiple tags, indicating the thematic similarity of questions within each tag.
To ensure the quality and diversity of our benchmark, 
we choose 23 different tags, including javascript, python, C++, \emph{etc.}, and collect top 2500 questions from each tag based on popularity.

\subsection{Test Case Building}
For both tasks, we construct test cases according to their token length (measured by GPT-4 tokenizer). 
We regard token lengths between 1,000 to 16,000 as \textbf{long-context} settings and text lengths exceed 16,000 as \textbf{ultra-long-context}  settings. 

Under long-context settings, TSort cases span test cases with 2k, 4k, 8k, and 16k tokens.
For each length, 
we fix the segment number N=4 and the length upper limit for each text segment and adjacent paragraphs before and after these contiguous chapters. 
We ensure that each text segment contains complete paragraphs thus no paragraph is sliced in the middle.
To build test cases with different contents, 
we set stride between beginning paragraphs of test cases during construction. 
After prepending the instructions, we further filter test cases that exceed the token upper bound.

For BestAnswer, we generate test cases with 1k, 2k, 4k, 6k, 8k, 12k, and 16k tokens under long-context settings.
Test cases contain the distractor answers under corresponding question and adaptable number of distractor answers from other similar questions under each length setting. 
To make evaluation results directly comparable across different length settings in long context scenarios, 
we ensure that the questions within the BestAnswer benchmark remain unchanged, 
regardless of the case length.
In BestAnswer, we define the most helpful answer as the answer explicitly accepted by the inquirer, and adopt it as the `groundtruth answer'\footnote{
We do not choose the answer with the highest number of votes, since the vote number can be influenced by factors such as the answer posting time and the identity of the respondent, in addition to its quality.}. 
For integrity reasons, we exclude all questions where the corresponding most helpful answer is not text-only. 
When choosing the distractors, we only consider answers that are provided prior to the accepted answer under corresponding question. 
Besides, we incorporate answers from other questions with similar tags to the original question to serve as distractor answers. 

Under ultra-long-context settings, 
we build test cases with 32k, 64k, and 128k tokens for both tasks. 
The construction paradigm is similar to the long-context setting. 
For BestAnswer, since the number of similar questions and the corresponding answers are limited, 
we relax tag similarity constraints and allow answers of questions with less similar tags to serve as the distractor answers.

\section{Evaluation Results}

\subsection{Experiment Setup}

We evaluate the following LLMs under long-context settings:
\napi proprietary models: (1) GPT-4-Turbo-0125, (2) GPT-4-Turbo-1106 (3) GPT-3.5-Turbo-1106, (4) Claude-2;
and \nhf open-source models: (5) LongChat-7b-v1.5-32k\citep{zheng2023judging}, (6) ChatGLM2-6B-32k\citep{zeng2022glm}, (7) ChatGLM3-6B-32k\citep{zeng2022glm}, (8) Vicuna-7b-v1.5-16k\citep{zheng2023judging}, (9) Vicuna-13b-v1.5-16k\citep{zheng2023judging}, (10) InternLM2-7b\citep{cai2024internlm2}.
Due to the inferior performance of open-source LLMs under long-context settings, 
only models with good performance (GPT-4-Turbo, Claude-2, \emph{etc.}) are evaluated under ultra-long-context settings.

For open-source LLMs, we sample a 1000-testcase subset for evaluation under each length setting. 
Due to the costly API of state-of-the-art proprietary models (GPT-4-Turbo, Claude-2, \emph{etc.}), 
we adopt 200-testcase subset (sampled from the 1000-testcase set) for evaluation under long-context settings, 
and a 50-testcase subset for evaluation under ultra-long-context settings.
All experiments are conducted using the open-source LLM evaluation platform OpenCompass~~\citep{2023opencompass}.
We adopt the zero-shot setting for all evaluation, 
and provide a `random guess' baseline. 
We also measure the instruction following rate and the copy instruction rate\footnote{Instruction following rate denotes if the LLM outputs follow the pre-defined format. Copy instruction rate measures if the LLM outputs the same answer as in-context example provides.} on both tasks.  

\subsection{Long-Context Evaluation Results}

\begin{table}[h]
    \centering
    \resizebox{1.0\linewidth}{!}{
        \tablestyle{6pt}{1.3}
        \begin{tabular}{c|c|c|c|c}
        \shline
        TSort  & 2k & 4k & 8k & 16k \\
        \shline
         GPT-4-Turbo-0125 & 15.5 & \textbf{16.5} & \textbf{8.5} & \textbf{5.5} \\
         GPT-4-Turbo-1106 & \textbf{18.5} & 15.5 & 7.5 & 3.5 \\
         GPT-3.5-Turbo-1106  & 4.0 & 4.5 & 4.5 & \textbf{5.5} \\
         Claude-2 & 5.0 & 5.0 & 4.5 & 3.0 \\ \shline
         LongChat-7b-v1.5-32k & 5.3 & 5.0 & 3.1 & 2.5 \\
         ChatGLM2-6B-32k & 0.9 & 0.7 & 0.2 & 0.9 \\
         ChatGLM3-6B-32k & 2.3 & 2.4 & 2.0 & 0.7 \\
         Vicuna-7b-v1.5-16k & 5.3 & 2.2 & 2.3 & 1.7 \\
         Vicuna-13b-v1.5-16k & 5.4 & 5.0 & 2.4 & 3.1 \\ 
         InternLM2-7b & 5.1 & 3.9 & 5.1 & 4.3 \\ \shline
         Random Guess & 4.2 & 4.2 & 4.2 & 4.2\\
        \shline
        \end{tabular}
    }
    \caption{\textbf{TSort} results under long-context settings. 
    We fix the number of segments $\mathbf{N}=4$ for TSort evaluation,
    thus random guess accuracy is roughly 4.2\% (1 / 24).}
    \label{tab:TSort}
\end{table}


\noindent
\textbf{TSort.} 
\Cref{tab:TSort} displays the test accuracy of various LLMs on the TSort task.
This evaluation underscores the complexity of TSort, 
highlighting its intricate nature that necessitates a comprehensive understanding and reasoning across long text.
Under settings from 2,000 to 8,000 tokens, 
only the most powerful proprietary model GPT-4-Turbo outputs the correct order of texts with a significant higher probability compared to the random baseline. 
When the context window expands to 16,000, the quality of GPT-4-Turbo's predictions also deteriorates to the random guess level. 
Other LLMs, encompassing both proprietary models and open-source models, all displaying similar performance compared to random guess (even under the relative short 2k setting). 
The results indicate that the TSort task posts a severe challenge to existing LLMs.

\begin{table*}[h]
    \centering
    \resizebox{.8\linewidth}{!}{
    \tablestyle{10pt}{1.3}
    \begin{tabular}{c|c|c|c|c|c|c|c}
    \shline
    BestAnswer & 1k & 2k & 4k & 6k & 8k & 12k & 16k \\
    \shline
         GPT-4-Turbo-0125 & 73.5 & \textbf{73.5} & 65.5 & \textbf{63.0} & \textbf{56.5} & \textbf{52.0} & \textbf{44.5} \\
         GPT-4-Turbo-1106 & \textbf{74.0} & \textbf{73.5} & \textbf{67.5} & 59.5 & 53.5 & 49.5 & 44.0 \\
         GPT-3.5-Turbo-1106  & 61.5 & 48.5 & 41.5 & 29.5 & 17.0 & 2.5 & 2.5 \\
         Claude-2 & 65.0 & 43.5 & 23.5 & 15.0 & 17.0 & 12.0 & 11.0 \\ \shline
         LongChat-7b-v1.5-32k & 32.4 & 10.7 & 5.7 & 3.1 & 1.9 & 1.6 & 0.8\\
         ChatGLM2-6B-32k & 31.2 & 10.9 & 4.5 & 1.6 & 1.6 & 0.0 & 0.3 \\
         ChatGLM3-6B-32k & 39.8 & 18.8 & 9.0 & 5.0 & 3.4 & 0.9 & 0.5 \\
         Vicuna-7b-v1.5-16k & 37.0 & 11.1 & 5.8 & 3.2 & 1.8 & 1.9 & 1.0 \\
         Vicuna-13b-v1.5-16k & 53.4 & 29.2 & 13.1 & 4.3 & 2.2 & 1.4 & 0.9 \\
         InternLM2-7b & 58.6 & 49.5 & 33.9 & 12.3 & 13.4 & 2.0 & 0.8 \\
         Random Guess & 26.7 & 10.1 & 4.5 & 3.0 & 2.3 & 1.4 & 1.1\\
        \shline
        \end{tabular}
    }
    \caption{\textbf{BestAnswer} results under long-context settings. For a question with $\mathbf{N}$ candidate answers, we define the random guess accuracy as $1 / \mathbf{N}$. The random guess accuracy over a long-context setting is the average of random guess accuracy for all questions within the test set. }
    \label{tab:BestAnswer}
\end{table*}

\noindent
\textbf{BestAnswer. } 
\Cref{tab:BestAnswer} presents the test accuracy of LLMs on BestAnswer.
GPT-4-Turbo establishes the state-of-the-art on the BestAnswer benchmark.
It achieves an outstanding 44.5\% accuracy under the 16k long-context setting, where around 100 distractor answers exist for each question. 
Among other proprietary models, Claude-2 achieves the second best accuracy 11\% under the 16k setting. 
GPT-3.5-Turbo-1106, while outperforming Claude-2 under some relative short settings (2k, 4k, 6k), demonstrates performance similar to random guess under the 16k setting. 
There is a considerable performance gap between proprietary models and open-source models on BestAnswer. 
Although some models like Vicuna-13b-v1.5-16k and InternLM2-7b perform well under short settings, 
a dramatic accuracy decline can be observed when text length becomes larger.

\begin{table}[h]
    \centering
    \resizebox{1.0\linewidth}{!}{
        \tablestyle{6pt}{1.3}
        \begin{tabular}{c|c|c|c|c}
        \shline
        CopyInst Rate & 2k & 4k & 8k & 16k \\
        \shline
         GPT-4-Turbo-1106 & 25.0 & 22.0 & 10.5 & 1.0 \\
         GPT-3.5-Turbo-1106  & 30.0 & 25.5 & 64.5 & 73.3 \\
         Claude-2 & 99.5 & 95.0 & 97.4 & 96.9 \\ \shline
         Expectation & 5.0 & 5.0 & 5.0 & 5.5 \\ \shline
         LongChat-7b-v1.5-32k & 100.0 & 99.8 & 99.1 & 100.0 \\
         ChatGLM2-6B-32k & 11.3 & 13.8 & 10.5 & 81.3 \\
         ChatGLM3-6B-32k & 21.6 & 54.8 & 88.0 & 88.1 \\
         Vicuna-7b-v1.5-16k & 100.0 & 100.0 & 59.4 & 33.3 \\
         Vicuna-13b-v1.5-16k & 96.6 & 99.0 & 12.2 & 3.1 \\ \shline
         Expectation & 5.3 & 5.0 & 5.4 & 5.2 \\ \shline
        \end{tabular}
    }
    \caption{The copy instruction rate of LLMs on \textbf{TSort} under long-context settings. \textbf{Expectation} means the ratio of test cases for which the in-context example answer is exactly the correct one. }
    \label{tab:textsort_copyinst}
\end{table}

\begin{table}[h]
    \centering
    \resizebox{\linewidth}{!}{
    \tablestyle{10pt}{1.3}
    \begin{tabular}{c|c|c|c|c|c|c|c}
    \shline
    CopyInst Rate & 1k & 2k & 4k & 6k & 8k & 12k & 16k \\
    \shline
         GPT-4-Turbo-1106 & 12.5 & 8.5 & 5.0 & 5.5 & 6.0 & 2.0 & 2.0 \\ 
         GPT-3.5-Turbo-1106  & 16.5 & 22.5 & 18.5 & 16.0 & 11.5 & 2.0 & 0.0 \\
         Claude-2 & 21.5 & 25.5 & 40.5 & 41.0 & 42.5 & 49.0 & 55.0 \\ \shline
         Expectation & 13.0 & 7.0 & 3.0 & 2.0 & 2.5 & 1.5 & 1.5\\ \shline
         LongChat-7b-v1.5-32k & 67.4 & 94.7 & 89.5 & 57.8 & 70.6 & 49.4 & 13.0\\
         ChatGLM2-6B-32k & 36.5 & 43.7 & 35.8 & 27.2 & 24.4 & 35.5 & 44.7 \\
         ChatGLM3-6B-32k & 47.9 & 66.1 & 33.3 & 30.4 & 22.5 & 24.8 & 16.7 \\
         Vicuna-7b-v1.5-16k & 63.1 & 96.2 & 91.8 & 57.9 & 66.6 & 27.8 & 17.9 \\
         Vicuna-13b-v1.5-16k & 27.8 & 45.8 & 55.3 & 19.8 & 3.4 & 5.6 & 11.1 \\ \shline
         Expectation & 14.4 & 10.0 & 5.1 & 2.3 & 1.7 & 1.3 & 1.2\\
        \shline
        \end{tabular}
    }
    \caption{The copy instruction rate of LLMs on \textbf{BestAnswer} under long-context settings. \textbf{Expectation} means the ratio of test cases for which the in-context example answer is exactly the correct one.}
    \label{tab:BestAnswer_copyinst}
\end{table}



\begin{figure*}[t]
    \centering
    \includegraphics[width=.95\linewidth]{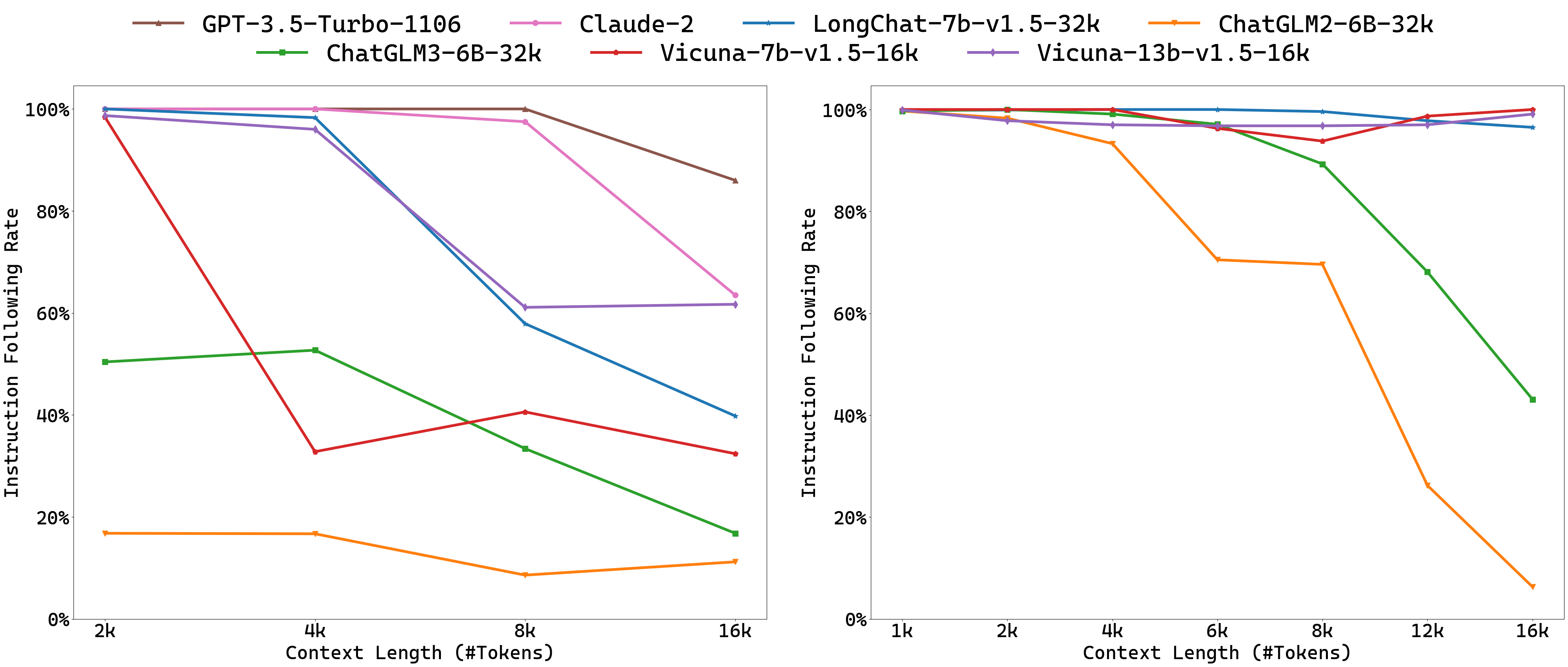}
    \caption{The instruction following rate of LLMs on TSort (Left) and BestAnswer (Right) under long-context settings.
    GPT-4-Turbo on TSort and all proprietary models on BestAnswer achieve 100\% instruction following rate across all long-context settings, 
    thus not displayed.}
    \label{fig:instr-follow}
\end{figure*}

\subsection{Error Breakdown}
We further analyze the error instances on TSort and BestAnswer, 
and find that most errors can be attributed to two categories: 
1. The LLM fails to follow the provided instruction and does not output a valid answer\footnote{A valid answer contains a permutation of N segment numbers on TSort and at least one designation of answers on BestAnswer.};
2. The LLM does output a valid answer. However, it simply copies the example answer we provide in the in-context example. 
\Cref{fig:instr-follow} display instruction following rate on TSort and BestAnswer.
\Cref{tab:textsort_copyinst,tab:BestAnswer_copyinst} provide detailed statistics about the copy instruction rate on TSort and BestAnswer.

The state-of-the-art GPT-4-Turbo maintains a relatively low copy instruction rate and impeccable instruction following rate on both tasks. 
Error instances of Claude-2, LongChat and Vicuna models are predominantly due to elevated Copy Instruction Rate, 
while ChatGLM models suffer from low instruction following rate. 
It is worth noting that all models, with the sole exception of GPT-4-Turbo, 
find it more difficult to follow the instruction on both tasks as text length increases.

\subsection{Ultra-Long-Context Evaluation Results}

We evaluate the following proprietary models under ultra-long-context settings. (1) GPT-4-Turbo-0125 (2) GPT-4-Turbo-1106 (3) Claude-2. (4) Claude-2.1.  We also evaluate InternLM2-7b on BestAnswer benchmark under ultra-long-context settings.
Due to high API calling expense, we test 50 samples under each ultra-long context setting. Table \ref{tab:BestAnswer_ultralong} demonstrates the result.

\begin{table}[h]
    \centering
    \resizebox{1.0\linewidth}{!}{
        \tablestyle{8pt}{1.3}
        \begin{tabular}{c|c|c|c|c}
        \shline
        Benchmark & Model & 32k & 64k & 128k \\
        \shline
        \multirow{4}{*}{TSort}
        & GPT-4-Turbo-0125 & 2.0 & 4.0 & 2.0 \\
        & GPT-4-Turbo-1106 & \textbf{6.0} & \textbf{6.0} & \textbf{6.0} \\
        & Claude-2 & 0.0 & 0.0 & / \\
        & Claude-2.1 & 0.0 & 0.0 & 0.0 \\ \scline{2-5}
        & Random Guess & 4.2 & 4.2 & 4.2 \\
        \shline
        \multirow{4}{*}{BestAnswer}
        & GPT-4-Turbo-0125 & \textbf{30.0} & 0.0 & 0.0 \\
        & GPT-4-Turbo-1106 & 16.0 & 0.0 & 0.0 \\
        & Claude-2 & 4.0 & 0.0 & / \\
        & Claude-2.1 & 4.0 & 0.0 & 0.0 \\
        & InternLM2-7b & 0.5 & \textbf{0.5} & 0.0 \\ \scline{2-5}
        & Random Guess & 0.6 & 0.3 & 0.1 \\
        \shline
        \end{tabular}
    }
    \caption{Results of LLMs on TSort and BestAnswer benchmarks in ultra-long context settings.}
    \label{tab:BestAnswer_ultralong}
\end{table}

Though the evaluated models claim that they can understand long text up to 100,000+ tokens (a whole book with hundreds of pages, \emph{e.g.}), 
they suffer from a dramatic decline on their performance under ultra-long-context settings, 
comparing to their long-context performance. 
For the TSort task, GPT-4-Turbo is able to achieve a random guess level accuracy,  
while Claude fails to give any correct answers. 
For BestAnswer, the performance of all three models fall sharply from 16k to 32k text length.
Meanwhile, they can not give any correct answer when the text length is greater than 32k.

\subsection{Ablation Study}

\subsubsection{Perplexity Evaluation on TSort}

Perplexity (PPL) evaluation is frequently adopted to assess the capability of LLMs. 
During inference, models compute the perplexity of multiple candidates and the one with the lowest perplexity is selected as the inference result. 
For TSort, we create 24 candidates for perplexity computation, 
each candidate is a permutation of the 4 text segments. 
We conduct PPL-based evaluation for open-source LLMs on 2k, 4k and 8k text length settings. 
\Cref{tab:TSort_ppl} exhibits the PPL-Eval result on TSort. 
When text segments are arranged in the correct order, 
a significantly lower perplexity score can usually be observed\footnote{
One potential cause is that the chapters have been used for pretraining.}, 
resulting in the high TSort accuracy. 
However, when the sorting task is presented as QAs where LLMs are asked to directly output the correct order, the performance significantly deteriorates,
indicating the limited instruction following capabilities of existing LLMs.

\begin{table}[h]
    \centering
    \resizebox{1.0\linewidth}{!}{
        \tablestyle{10pt}{1.3}
        \begin{tabular}{c|c|c|c}
        \shline
        TSort (PPL Eval) & 2k & 4k & 8k \\
        \shline
         LongChat-7b-v1.5-32k & 60.9 & 68.3 & 77.4 \\
         ChatGLM2-6B-32k & 40.5 & 53.5 & 57.5 \\
         ChatGLM3-6B-32k & 50.1 & 57.0 & 59.3 \\
         Vicuna-7b-v1.5-16k & 70.1 & 78.3 & 77.7 \\
         Vicuna-13b-v1.5-16k & 79.3 & 86.7 & 89.2 \\  \shline
         Random Guess & 4.2 & 4.2 & 4.2 \\
        \shline
        \end{tabular}
    }
    \caption{Perplexity Evaluation Results on \textbf{TSort} for open-source LLMs.}
    \label{tab:TSort_ppl}
\end{table}


\subsubsection{Position Bias in BestAnswer}

To study the position bias of existing LLMs, 
in BestAnswer, 
we keep questions and answer candidates the same and alter the position of groundtruth answers.
Specifically, we manually set the groundtruth answer at the beginning, in the middle, or at the rear of all answers and then perform the evaluation. 
\Cref{tab:BestAnswer_answerpos} displays the evaluation results.
\textbf{All models demonstrate significant position bias in choosing the most helpful answer.} 
Most models achieve much better accuracy when the most helpful answer presents at the beginning. 
Claude-2 has some unique behaviors. 
It performs the best when the groundtruth is positioned at the rear across 4 of 5 different settings.
As the input length increases, the position bias becomes more obvious. 
For instance, Vicuna-7b-v1.5-16k demonstrates relatively uniform accuracy under the 1k setting. 
However, when the input length extends to 16k tokens, 
the model's performance remains stable only when the best answer is at the front. 


\begin{table}[h]
    \centering
    \resizebox{1.0\linewidth}{!}{
        \tablestyle{5pt}{1.2}
        \begin{tabular}{c|c|c|c|c|c|c}
        \hline
        BestAnswer & Pos & 1k & 2k & 4k & 8k & 16k \\
        \hline
        \multirow{3}{*}{GPT-4-Turbo-1106}
        & front & 76.5 & 82.5 & 86.5 & 90.0 & 82.0 \\
        & mid & 74.5 & 68.0 & 60.0 & 38.0 & 38.5 \\
        & rear & 57.5 & 46.6 & 44.0 & 40.5 & 26.5 \\
        \hline
        \multirow{3}{*}{GPT-3.5-Turbo-1106}
        & front & 77.0 & 80.5 & 77.0 & 46.5 & 2.5 \\
        & mid & 64.5 & 48.5 & 32.0 & 9.5 & 0.5 \\
        & rear & 37.5 & 19.0 & 8.5 & 6.0 & 3.5 \\
        \hline
        \multirow{3}{*}{Claude-2}
        & front & 34.0 & 19.0 & 14.5 & 50.0 & 6.0 \\
        & mid & 49.0 & 35.5 & 21.5 & 13.0 & 5.0 \\
        & rear & 59.0 & 36.5 & 26.0 & 11.0 & 9.5 \\
        \hline
        \multirow{3}{*}{LongChat-7b-v1.5-32k}
        & front & 24.1 & 5.0 & 12.1 & 33.6 & 29.0 \\
        & mid & 32.7 & 13.6 & 0.2 & 0.2 & 0.0 \\
        & rear & 29.8 & 1.9 & 0.0 & 0.1 & 0.1 \\
        \hline
        \multirow{3}{*}{ChatGLM2-6B-32k}
        & front & 30.0 & 31.5 & 46.2 & 10.5 & 0.5 \\
        & mid & 27.7 & 10.4 & 1.0 & 0.1 & 0.1 \\
        & rear & 28.5 & 12.4 & 2.6 & 4.1 & 0.0 \\
        \hline
        \multirow{3}{*}{ChatGLM3-6B-32k}
        & front & 48.9 & 34.3 & 37.6 & 35.8 & 19.0 \\
        & mid & 41.9 & 22.3 & 5.3 & 0.9 & 0.1 \\
        & rear & 28.8 & 5.4 & 3.7 & 8.8 & 2.9 \\
        \hline
        \multirow{3}{*}{Vicuna-7b-v1.5-16k}
        & front & 29.3 & 8.9 & 14.0 & 37.6 & 25.4 \\
        & mid & 32.8 & 13.6 & 0.0 & 0.0 & 0.2 \\
        & rear & 34.2 & 2.1 & 0.0 & 0.0 & 0.7 \\
        \hline
        \multirow{3}{*}{Vicuna-13b-v1.5-16k}
        & front & 52.5 & 51.4 & 58.6 & 81.7 & 11.8 \\
        & mid & 64.5 & 29.2 & 1.5 & 0.5 & 0.3 \\
        & rear & 34.2 & 2.4 & 0.0 & 0.0 & 13.4 \\
        \hline
        \end{tabular}
    }
    \caption{Results of LLMs on \textbf{BestAnswer} where the best answer is set at the front, in the middle and at the rear of all answers. 
    Pos denotes the position of the best answer.}
    \label{tab:BestAnswer_answerpos}
\end{table}

\subsubsection{Scalable Position Embeddings}

Scalable position embeddings have shown their value in extending context window while requiring minimal or no fine-tuning steps. 
Existing position embedding methods for context window extension can be categorized into 
two major categories: position interpolation and length extrapolation.
NTK-aware Scaled RoPE utilizes the advantage of both methods by changing the base of RoPE. 
ReRoPE and Leaky ReRoPE~\citep{rerope2023} design a window size to control the application of scalable position embeddings directly. 
We conduct our study on Vicuna-v1.5 models~\citep{zheng2023judging}, 
which are Llama 2 fine-tuned with 4k context window. 
We adopt original models (4k context window) as the baseline across all settings. 
\Cref{tab:extrapolation} shows the result of different position embedding methods on the BestAnswer benchmark.
Our findings indicate that \textbf{scalable position embeddings do improve the long-context modeling capability}.
All methods enhance the accuracy under the 8k setting, which is beyond the original context window.
Concurrently, the model performance under short settings (1k, \emph{e.g.}) is basically retained. 
NTK-aware Scaled RoPE diminishes performance on 1k context length, 
but outperforms other two methods on longer context. 
The advantage of these methods is more obvious on Vicuna-13b-v1.5. 
Moreover, comparing to their 16k versions, 
which utilize Flash Attention and are further trained on high-quality 16k length conversation data, 
advanced scalable position embeddings still achieve comparable performance.

\begin{table}[ht]
    \centering
    \resizebox{1.0\linewidth}{!}{
        \begin{tabular}{c|c|c|c|c}
        \hline
        Vicuna-7b-v1.5 & 1k & 2k & 4k & 8k \\
        \hline
        ReRoPE & 39.6/39.6 & 11.6/11.6 & 4.7/5.4 & 2.3/3.2 \\
        Leaky ReRoPE & 39.9/39.9 & 11.2/11.2 & 5.1/5.7 & 1.3/2.0 \\
        NTK & 32.5/32.5 & 10.7/10.7 & 5.8/5.8 & 3.9/3.9 \\
        Original(4k) & 39.5/39.5 & 9.8/11.0 & 4.2/5.5 & 0.0/0.0 \\
        Original(16k) & 37.0/39.5 & 11.1/11.1 & 5.8/5.8 & 2.5/2.7 \\
        \hline
        Vicuna-13b-v1.5 & 1k & 2k & 4k & 8k \\
        \hline
        ReRoPE & 49.2/49.2 & 22.5/22.5 & 9.2/10.0 & 1.5/2.8 \\
        Leaky ReRoPE & 49.3/49.3 & 23.8/23.8 & 8.7/9.8 & 1.3/2.6 \\
        NTK & 43.8/43.8 & 23.0/23.0 & 11.1/11.1 & 2.3/2.3 \\
        Original(4k) & 49.1/49.1 & 17.7/17.7 & 5.9/5.9 & 0.1/1.0 \\
        Original(16k) & 53.4/53.4 & 29.2/29.2 & 13.1/13.5 & 2.6/2.7 \\
        \hline
        \end{tabular}
    }
    \caption{Results of Vicuna-v1.5 with different context window extrapolation methods on \textbf{BestAnswer}. 
    `Original (4k) / (16k)' denotes the original Vicuna model trained with 4k / 16k context lengths. 
    In the reported `X/Y', X indicates the accuracy while Y indicates the accuracy which cases failed to follow the instruction are excluded.}
    \label{tab:extrapolation}
\end{table}

\subsubsection{Comparison with Other Long-Context Benchmarks}

We compare Ada-LEval with other long-context benchmarks to validate that our benchmarks require much overall text understanding to complete the task than traditional long-context benchmarks.

We regard a task requires models to understand text comprehensively if the performances of models decrease sharply when the text is truncated. 
TSort task meets this requirement since truncating any segment will lead to an incorrect answer.

To exhibit the BestAnswer requires more comprehensive text understanding than traditional QA and summarization tasks, we conduct an experiment on BestAnswer(16k version) and 2 classic long-context datasets, NarrativeQA(LongBench subset, QA task) and GovReport(LongBench subset, summarization task) respectively. 
The metric for NarrativeQA is F1 score and metric for GovReport is Rouge-L. We evaluate the performance of GPT-4-Turbo-1106 on all 3 datasets. Each test case is truncated into 2k, 4k and 8k version as the input. We also provide its full version for comparison.

\begin{table}[h]
    \centering
    \resizebox{1.0\linewidth}{!}{
        \tablestyle{8pt}{1.3}
        \begin{tabular}{c|c|c|c|c|c}
        \shline
        Benchmark & 2k & 4k & 8k & Full & Avg \#tokens \\
        \shline
        BestAnswer & 11.0 & 20.0 & 31.5 & 44.0 & 15646 \\
        NarrativeQA & 24.7 & 25.6 & 29.7 & 33.1 & 10276 \\
        GovReport & 30.7 & 32.4 & 33.6 & 30.9 & 29872 \\
        \shline
        \end{tabular}
    }
    \caption{Results of GPT-4-Turbo on different long-context benchmarks.}
    \label{tab:different_benchmarks}
\end{table}

From the table \ref{tab:different_benchmarks}, the performance of GPT-4-Turbo on BestAnswer decreases more dramatically than NarrativeQA and GovReport when text is truncated. Notably, the performance on GovReport even increases when text is truncated into 4k and 8k. Therefore, our benchmarks require more full-text comprehension than traditional QA and summarization tasks.
\section{Conclusion}
In this paper, we introduce Ada-LEval, a length-adaptable dataset to assess long-context capability of LLMs. 
We conduct comprehensive experiments on multiple LLMs and find that all open-source models still lag significantly behind state-of-the-art proprietary models in terms of long context capability. 
When the input length scales to 4,000 tokens, 
most open-source models rapidly deteriorates to random guess level. 
In the meanwhile, the capability of proprietary models is also severely limited,
When it comes to the ultra-long setting (32,000+ tokens), no proprietary model notably outperforms the random baseline. 
Ada-LEval is the first benchmark that evaluates LLMs under the ultra-long setting, 
and we hope that the limitations pointed out by this benchmarks can serve as valuable references for future developments of long-context LLMs.

\noindent \textbf{Acknowledgement. }
This project is supported by the 
National Key R\&D Program of China No.2022ZD0161600 and the
Shanghai Postdoctoral Excellence Program (No.2023023).

\section{Limitations}

Ada-LEval is a challenging benchmark, requiring strong understanding and reasoning capabilities over long text. 
Due to the poor instruction following rate and copy instruction rate of open-source LLMs, Ada-LEval can hardly distinguish their long context capability through the accuracy metric.

Furthermore, as text length increases, the difficulty of Ada-LEval rises sharply under ultra-long-context settings. Even state-of-the-art proprietary models are not able to achieve an ideal performance, which further constrains its applicability to current LLMs.

\bibliography{main}
\newpage
\appendix
\section{Test Case Building Statistics}

Recall that for each case length on Tsort task, we set the length upper limit for each text segment and the neighboring paragraphs before and after these contiguous chapters. We also set stride between beginning paragraphs. \Cref{tab:textsort_build} demonstrates the detail statistics on the length upper limit and the stride.

\noindent
On BestAnswer task, two questions are regarded as similar questions when they have 40\% tags in common. Under ultra-long-context settings, both questions should contain at least 1 tag in common.

\section{Evaluation Setups}
\label{sec:appendix}

\textbf{Evaluation Hyperparameters.} For open-source LLMs, we adopt their default hyperparameters during evaluation on Ada-LEval.  
For proprietary models including GPT-4-Turbo, GPT-3.5-Turbo-1106, we set the temperature to 0.

\noindent
\textbf{Computational Budget.} Our experiments for open-source LLMs are conducted on NVIDIA A100 80GB GPU. 
The entire evaluation consumes around 800 GPU-hours.

\noindent
\textbf{Benchmark Instructions.} 
We present instructions of both tasks within Ada-LEval. 
To ensure that models know what to do, we contain the sample input and output format that models need to follow in solving problems. 
The instructions are shown below.

\noindent
\textbf{Validity of 200-testcase subset.} Our experiments on long-context settings adopt 200-testcase subset for proprietary models and 1000-testcase subset for open-source LLMs. To ensure that evaluation results on 200-testcase subset is valid, \Cref{tab:textsort_200} and \Cref{tab:BestAnswer_200} display results on 200-testcase subset.

\begin{table}[ht]
    \centering
    \tablestyle{10pt}{1.3}
    \resizebox{1.0\linewidth}{!}{
        \begin{tabular}{c|c|c|c|c}
        \hline
        Setting & Before & Segments & After & Stride \\
        \hline
         2k & 200 & 350 & 200 & 64 \\
         4k  & 300 & 800 & 300 & 64 \\
         8k & 400 & 1750 & 400 & 64 \\
         16k & 500 & 3700 & 500 & 64 \\
         32k & 500 & 7700 & 500 & 128 \\
         64k & 500 & 15700 & 500 & 128 \\
         128k & 500 & 31700 & 500 & 128 \\
        \hline
        \end{tabular}
    }
    \caption{The length upper limit of text segments and stride between beginning paragraphs on TSort.}
    \label{tab:textsort_build}
\end{table}

\begin{table}[h]
    \centering
    \tablestyle{10pt}{1.3}
    \resizebox{1.0\linewidth}{!}{
        \begin{tabular}{c|c|c|c|c}
        \hline
        TSort (200-testcase) & 2k & 4k & 8k & 16k \\
        \hline
         GPT-4-Turbo-0125 & 15.5 & \textbf{16.5} & \textbf{8.5} & \textbf{5.5} \\
         GPT-4-Turbo-1106 & \textbf{18.5} & 15.5 & 7.5 & 3.5 \\
         GPT-3.5-Turbo-1106  & 4.0 & 4.5 & 4.5 & \textbf{5.5} \\
         Claude-2 & 5.0 & 5.0 & 4.5 & 3.0 \\
         LongChat-7b-v1.5-32k & 5.0 & 5.0 & 2.5 & 2.0 \\
         ChatGLM2-6B-32k & 1.0 & 0.5 & 0.5 & 1.0 \\
         ChatGLM3-6B-32k & 3.5 & 3.0 & 1.0 & 0.5 \\
         Vicuna-7b-v1.5-16k & 5.0 & 1.5 & 1.0 & 2.5 \\
         Vicuna-13b-v1.5-16k & 5.0 & 5.0 & 3.0 & 4.0 \\
         Random Guess & 4.2 & 4.2 & 4.2 & 4.2\\
        \hline
        \end{tabular}
    }
    \caption{\textbf{TSort} results under long-context settings(200-testcase subset).}
    \label{tab:textsort_200}
\end{table}

\begin{table*}[h]
    \centering
    \resizebox{.8\linewidth}{!}{
    \tablestyle{10pt}{1.3}
    \begin{tabular}{c|c|c|c|c|c|c|c}
    \hline
    BestAnswer (200-testcase) & 1k & 2k & 4k & 6k & 8k & 12k & 16k \\
    \hline
         GPT-4-Turbo-0125 & 73.5 & \textbf{73.5} & 65.5 & \textbf{63.0} & \textbf{56.5} & \textbf{52.0} & \textbf{44.5} \\ 
         GPT-4-Turbo-1106 & \textbf{74.0} & \textbf{73.5} & \textbf{67.5} & 59.5 & 53.5 & 49.5 & 44.0 \\
         GPT-3.5-turbo-1106  & 61.5 & 48.5 & 41.5 & 29.5 & 17.0 & 2.5 & 2.5 \\
         Claude-2 & 65.0 & 43.5 & 23.5 & 15.0 & 17.0 & 12.0 & 11.0 \\
         LongChat-7b-v1.5-32k & 32.5 & 8.0 & 3.5 & 3.0 & 2.5 & 1.5 & 1.0\\
         ChatGLM2-6B-32k & 36.0 & 10.5 & 3.0 & 0.5 & 1.5 & 0.0 & 0.0 \\
         ChatGLM3-6B-32k & 37.0 & 15.5 & 5.5 & 4.0 & 5.5 & 0.5 & 0.5 \\
         Vicuna-7b-v1.5-16k & 32.5 & 8.5 & 2.5 & 3.5 & 3.0 & 0.5 & 2.0 \\
         Vicuna-13b-v1.5-16k & 52.0 & 29.0 & 11.0 & 4.0 & 1.5 & 1.0 & 1.5 \\
         Random Guess & 26.7 & 10.1 & 4.5 & 3.0 & 2.3 & 1.4 & 1.1\\
        \hline
        \end{tabular}
    }
    \caption{\textbf{BestAnswer} results under long-context settings(200-testcase subset). }
    \label{tab:BestAnswer_200}
\end{table*}

\begin{tcolorbox}[colback=gray!10,
                  colframe=black,
                  width=\linewidth,
                  arc=3mm, auto outer arc,
                 ]
  \textbf{TSort:}
  
  You are an AI assistant. Your job is to sort multiple book sections into the correct order. 
  
  Each time, you will be provided with 4 pieces of text.

  These texts form a continuous part of a book, but are provided in random order.
  
  You need to find the correct order and return the answer in a string.
  
  For example, if you output [4, 1, 3, 2], that means the correct order is: Part 4 -> Part 1 -> Part 3 -> Part 2.
    
  You will also be provided with the neighboring paragraphs before and after the 4 pieces of texts.
  
  The case sample is shown below and you should give me the answer in the format exactly the same as the sample.
  
  However, you should NOT focus on the content of sample answer.
  
  Please do NOT output any extra content. 
  
  Sample Input (format only):

  Before: XXX (Text before the continuous book part)

  Part 1: XXX
  
  Part 2: XXX
    
  Part 3: XXX
  
  Part 4: XXX
  
  After: XXX (Text after the continuous book part)
  
  Sample Output (format only):
  
  Answer: [4, 1, 3, 2]
\end{tcolorbox}

\begin{tcolorbox}[colback=gray!10,
                  colframe=black,
                  width=\linewidth,
                  arc=3mm, auto outer arc,
                 ]
  \textbf{BestAnswer:}

    You are an AI assistant. Your job is to find out the most helpful answer to a given question.
    
    Each time, you will be provided with a question and n answers to this question.
    
    Each answer begins with an 'A' and a number(e.g. A4), which represents its designation.
    
    You need to determine which answer is the most helpful one to the question.
    
    The case sample is shown below and you should give me the answer in the format exactly the same as the sample.
    
    However, you should NOT focus on the content of sample answer.
    
    Sample Input (format only):
    
    The question is given below.
    
    XXX(The content of question)
    
    Possible answers are given below.
    
    A1:
    
    XXX(The content of answer 1)
    
    A2:
    
    XXX(The content of answer 2)
    
    .
    
    .
    
    .
    
    An:
    
    XXX(The content of answer n)
    
    Now the answers are over, please decide which answer is the most helpful one to the question. You must give me only the designation of the MOST helpful answer.
    
    Sample Output (format only):
    
    Answer: The designation of the most helpful answer.(e.g. A4 means answer 4 is the most helpful answer)

\end{tcolorbox}

\end{document}